\documentclass[11pt,a4paper]{article}
\usepackage[hyperref]{acl2021}
\usepackage{times}
\usepackage{latexsym}
\usepackage{graphicx}
\usepackage{float}
\usepackage{amsmath}

\usepackage{hyperref}
\hypersetup{
    colorlinks=true,
    linkcolor=magenta,
    filecolor=magenta,      
    urlcolor=cyan,
    citecolor=blue,
}

\usepackage{algorithm,algpseudocode}

\usepackage{microtype}
\aclfinalcopy

\title{New Methods \& Metrics for LFQA tasks}

\author{Suchismit Mahapatra \\
  \texttt{suchismi@buffalo.edu} \\
  \And
  Vladimir Blagojevic \\
  \texttt{dovlex@gmail.com} \\
  \AND
  Pablo Bertorello \\
  \texttt{edu@pablotech.uno} \\
  \And
  Prasanna Kumar \\
  \texttt{strategypk@gmail.com} \\}

\begin{document}
\maketitle

\begin{abstract}
Long-form question answering (LFQA) tasks require retrieving the documents pertinent to a query, using them to form a paragraph-length answer. Despite considerable progress in LFQA modeling, fundamental issues impede its progress: i) train/validation/test dataset overlap, ii) absence of automatic metrics and iii) generated answers not being ``grounded'' in retrieved documents. This work addresses every one these critical bottlenecks, contributing natural language inference/generation (NLI/NLG) methods and metrics that make significant strides to their alleviation.
\end{abstract}

\section{Introduction}\label{sec:introduction}

Most existing question answering (QA) algorithms struggle to provide rich explanations the way humans can. Sometimes systems are specific a knowledge domain, and in others they provide a single-word or single-phrase answer from a given input passage. Often the solution identifies a simple fact in a single passage or document, which is then presented as the answer to the formulated questions. 

Enter long-form question answering (LFQA), which remains a fundamental challenge in natural language processing (NLP). In general, LFQA tasks are difficult because they involve first to retrieve documents/passages that are relevant to a given question, subsequently using a text generation component to produce paragraph-length answers from these sources. Even though significant progress in LFQA has been made, using novel models/pipelines to improve retrieval and/or text generation components, much still remains unaddressed. The details of these challenges are elaborated in the next section.

\subsection{Issues in LFQA tasks}\label{ssec:lfqa_issues}
\citep{krishna2021} in their work discuss questionable trends in recent approaches to solve the LFQA task. Specifically, they highlight how the LFQA task formulation itself results in problematic evaluation and dataset creation. To corroborate their argument, they design a system that achieves state-of-the-art results on the ELI5 LFQA dataset~\citep{fan2019}. Subsequently, they critique their system to demonstrate glaring trends: i) the answers generated by their system are often not actually grounded in the retrieved documents, ii) there is a significant train/validation dataset overlap, iii) the ROUGE-L metric~\citep{lin2004} is not informative and can be easily gamed, and iv) human evaluation is unreliable for LFQA tasks.

Subsequently~\citep{krishna2021} discuss current open problems in LFQA tasks, namely: i) being able to automatically test and quantify ``data leakage'', that is train/validation/test dataset overlap, ii) the need for automatic metrics for LFQA tasks, and iii) preventing ``hallucination'', that is to generate answers that are grounded in the retrieved documents. Each of these needed enhancements is non-trivial, and highly significant towards solving the LFQA task as a whole.

%\subsubsection{Hallucination}\label{sssec:hallucination}
{\em Hallucination} refers to the factual inconsistency between source document and generated text. This remains a major issue both in tasks that summarize text, and in QA tasks. Refer to~\cite{xie2021, rebuffel2021} for additional discussion on potential causes and recent mitigation work.

\subsection{Contributions}\label{ssec:contributions}
Section~\ref{ssec:lfqa_issues} discusses the most prevalent issues in LFQA. The challenge is compounded by the scarcity of appropriate datasets. To date ELI5~\citep{fan2019} is the only publicly-available large-scale LFQA dataset. Problematically, this dataset has significant train/validation/test overlap (as much as 81\% of questions in the validation set occur in paraphrased form in the training set). Therefore, any LFQA model trained on ELI5 is suspect. 

The principal contributions of this work are:
\begin{itemize}
    \item{Validation of the ``data leakage'' in the ELI5 dataset, quantifying a new version with significantly less train/validation/test overlap.}
    \item{A framework/tool that may remove data overlap automatically, which continues in active development.}
    \item{The first known automatic metrics for LFQA tasks, providing better grounding for generated answers.}
\end{itemize}

The rest of the paper is organized as follows: related works are discussed in Section~\ref{sec:related_work}. Section~\ref{sec:methodology} proposes a novel metric and discusses how it is utilized for solving the above issues. Experimental results on benchmark datasets, are summarized in Section~\ref{sec:results}. Conclusions and discussion are in Section~\ref{sec:conclusion}.

\section{Related work}\label{sec:related_work}
This section provides an overview of different related works. In Sections~\ref{ssec:metrics} and~\ref{ssec:eval}, compares the prevalent metrics, examining issues that limit their applicability. This motivates the need for automatic metrics for QA/LFQA tasks. CoCo~\citep{xie2021} is explored in detail in Section~\ref{ssec:coco}, given its significance to the stated contributions. A brief overview of other relevant work is in Sections~\ref{ssec:other_automatic_metrics} and~\ref{ssec:retrieval_scoring_models}.

\subsection{Evaluation metrics}\label{ssec:metrics}
This section provides a brief overview of the prevalent {\em n}-gram based metrics to automatically evaluate machine generated text. See ~\cite{wang2020, zhang2019} for further discussion of these methods. Additionally, ~\citep{xie2021} discusses non {\em n}-gram based metrics.

ROUGE~\citep{lin2004} was one of the earlier metrics for tasks that summarize text. ROUGE-{\em n} and ROUGE-{\em L} are most commonly used variants. In the former, the value {\em n} is typically set to $\{1, 2\}$ and ROUGE-{\em n} computes the F1 score for all reference {\em n}-grams in the generated summary whereas in the latter {\em L} refers to the length of the longest common sub-sequence between the generated summary and references texts used.

Both BLEU~\cite{papineni2002} and METEOR~\cite{lavie2007} are closely related to ROUGE. Whereas BLEU was primarily intended for machine translation tasks and computes the precision of reference {\em n}-grams in the generated summary, METEOR~\cite{lavie2007} added more flexibility to BLEU by stemming and synonym replacement, in addition to text alignment, which results in more accurate scores in many scenarios.

Section~\ref{ssec:eval} discusses the shortcomings of metrics that are based on measuring {\em n}-gram based matching, motivating the need for automatic metrics for QA/LFQA tasks.

\subsection{Evaluation \& Motivation for automatic metrics}\label{ssec:eval}
Existing techniques for measuring quality in QA tasks are primarily based on counting {\em n}-gram overlap. These metrics depend on access to reference texts, enabling them to score text summary precision and recall, comparing the {\em n}-grams present in the summary to the ones in the reference.

Two main issues arise when using metrics which are {\em n}-gram based metrics. Firstly, they require one or more reference texts to compare against, which can be expensive and difficult to obtain. Additionally, many datasets contain only a single reference, which is not well suited for LFQA. Secondly, {\em n}-grams based approaches incorrectly place equal weights to all {\em n}-grams being matched, even as very few {\em n}-grams usually carry most of the information. This leads to insensitivity towards semantic errors. Moreover, prior work indicates that these metrics are poorly correlated with human judgments on factual consistency.

The above are the main reasons why human evaluation remains the primary method for evaluating quality of generated text. Additionally, manual evaluation is often slow and costly, major usage deterrents. Thus automatic metrics are necessary, which can without human intervention, accurately measure quality system performance on QA/LFQA tasks.

\subsection{CoCo}\label{ssec:coco}
\begin{figure}[!htbp]
\centering
\includegraphics[scale=0.5]{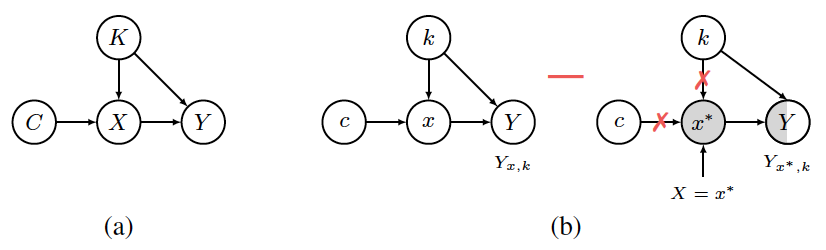}
\caption{Part (a) demonstrates the causal graph for text summarization, highlighting the relationships among the fact $\mathcal{\boldsymbol {C}}$, source document $\mathcal{\boldsymbol {X}}$, language prior $\mathcal{\boldsymbol {K}}$, and the model generated summary $\mathcal{\boldsymbol {Y}}$. The authors measure the causal effect of $\mathcal{\boldsymbol {X}}$ on $\mathcal{\boldsymbol {Y}}$ via subtracting the effect of $\mathcal{\boldsymbol {K}}$ on $\mathcal{\boldsymbol {Y}}$ from the total effect.}
\label{fig:coco}
\end{figure}

In CoCo~\citep{xie2021} the authors focus on the problem of factual inconsistency in generated summaries. They came up with an automatic evaluation metric which achieves the dual goals of improving correlation with human based judgments, while being convenient to use. The authors evaluate the factual consistency in the generated summaries via counterfactual estimation and remove the impact of language prior (refer to Figure~\ref{fig:coco} for details\footnote{Image taken from~\citep{xie2021}.}), which is a potential cause for factual inconsistency. Empirical evidence supports CoCo's efficacy. 

Section~\ref{sec:methodology} discusses a proposed adaptation of the CoCo metric for LFQA tasks. Additionally, CoCo is enhanced to understand how factually grounded the generated answer is to the retrieved documents. Algorithm~\ref{alg:coco_summarization} exemplifies the novel application of CoCo metric. 

\begin{algorithm}[H]
  \caption{CoCo metric for summarization tasks}
  \label{alg:coco_summarization}
  \begin{algorithmic}[1]
	\Require Source document $\mathcal{\boldsymbol {X}}$, model-generated summary $\mathcal{\boldsymbol {Y}}$, scoring model $\mathcal{\boldsymbol {F}}$
	\Ensure  CoCo score $\mathcal{\boldsymbol {C}}$ of $\mathcal{\boldsymbol {Y}}$ w.r.t. $\mathcal{\boldsymbol {X}}$
	\State Select key tokens $\mathcal{\boldsymbol {Y}}^\prime$ from $\mathcal{\boldsymbol {Y}}$.
	\State Mask the source document $\mathcal{\boldsymbol {X}}$ according $\mathcal{\boldsymbol {Y}}^\prime$ to produce $\mathcal{\boldsymbol {X}}^\prime$.
	\State Feed $\mathcal{\boldsymbol {X}}$ and $\mathcal{\boldsymbol {X}}^\prime$ into scoring model $\mathcal{\boldsymbol {F}}$ to generate token probabilities, i.e., $P (\boldsymbol{y}_i | \mathcal{\boldsymbol {X}}, \boldsymbol{y}_{< i})$ and $P (\boldsymbol{y}_i | \mathcal{\boldsymbol {X}}^\prime, \boldsymbol{y}_{< i})$, $\forall \boldsymbol{y}_i \in \mathcal{\boldsymbol {Y}}^\prime$.
	\State Calculate $\mathcal{\boldsymbol {C}} = \frac{1}{| \mathcal{\boldsymbol {Y}}^\prime |} \sum\limits_{\boldsymbol{y}_t \in \mathcal{\boldsymbol {Y}}^\prime} P (\boldsymbol{y}_t | \mathcal{\boldsymbol {X}}, \boldsymbol{y}_{< t}) - P (\boldsymbol{y}_t | \mathcal{\boldsymbol {X}}^\prime, \boldsymbol{y}_{< t})$
    \end{algorithmic}
\end{algorithm}

\subsection{Automatic metrics from other NLI/NLG tasks}\label{ssec:other_automatic_metrics}

QAGS~\citep{wang2020} focuses on summarization tasks, attempting to identify factual inconsistencies in the generated summaries. Internally, it uses a question answering (QA) model as well as a question generation (QG) model. It first uses the QG model to generate a set of questions about the summary.  Subsequently, it utilizes the QA model to answer these questions given the original text and summary independently. Finally it compares the similarity of corresponding answers to compute a quality score of how factual the summary is with regards to the original text.

BERTScore~\citep{zhang2019} measures sentence similarity using BERT~\citep{devlin2018} based contextual embeddings. It can be used as an automatic evaluation metric for natural language generation (NLG) tasks. BERTScore computes sentence similarity as a sum of cosine similarities between the sentences token embeddings. This is significant since the resulting scores correlate well to human evaluations, when compared to competing approaches.

QuestEval~\citep{scialom2021a} is an unified framework that can be used to evaluate generated summaries in NLG tasks. Unlike other approaches, it does not require any ground truth reference. This is significant since references are unavailable in many real-world scenarios. Its scores are substantially useful in terms of correlation with human judgments, when considering consistency, coherence, fluency, and relevance dimensions. QuestEval unifies  precision and recall-based QA metrics, resulting in a more robust metric, a major factor to success.

\subsection{Retrieval/Scoring models}\label{ssec:retrieval_scoring_models}

Models used to solve the LFQA task can be broadly divided into: i) {\em Extractive} models, which produce answers verbatim from the retrieved documents, and ii) {\em Abstractive} models, which can merge/regenerate the information content in the retrieved documents as necessary.

One of the main contributions of this work is the design of novel automatic metrics for LFQA tasks. We achieved this via adapting the CoCo~\citep{xie2021} metric, which was originally intended for summarization tasks, to instead use it for LFQA. Consequently, we experimented with different retrieval and scoring models included in~\citep{xie2021}, as well as different {\em Extractive} and {\em Abstractive} models typically used in practice. A brief overview of these different models follows.

BART~\citep{lewis2019} is modeled as a denoising autoencoder, pre-trained combining Bidirectional and Auto-Regressive
Transformer~\citep{vaswani2017} components. BART uses a neural machine translation architecture that, despite its simplicity, may generalize better than BERT~\citep{devlin2018} (uses a bidirectional encoder), and GPT~\citep{radford2019} (uses a left-to-right decoder). BART uses a variety of noising approaches, making it possible to learn a robust neural representations. It achieves state-of-the-art results on a range of abstractive dialogue, QA, and summarization tasks.

Sentence-BERT and SRoBERTa~\citep{reimers2019} are modifications of BERT~\citep{devlin2018} and RoBERTa~\citep{liu2019} respectively. They utilize siamese and triplet network topology to generate sentence embeddings, which are semantically meaningful. These adaptations improve the speed and efficiency of model training/inference. Importantly, Sentence-BERT and SRoBERTa embedings are amenable towards cosine similarity measures, while  BERT embeddings generally are not.

Bidirectional Encoder Representations from Transformers (BERT)~\citep{devlin2018} is designed to pre-train deep bidirectional representations from unlabeled text, by jointly conditioning on both left and right context. BERT incorporates two steps: i) pre-training and ii) fine-tuning. During pre-training, BERT is trained on unlabeled data over different tasks. For fine-tuning, the BERT model is first initialized with the pre-trained parameters, and all of the parameters are fine-tuned using labeled data from the downstream tasks. Each downstream task has separate fine-tuned models, even though they are initialized with the same pre-trained parameters. The pre-trained BERT model achieved state-of-the-art performance for a wide range of tasks, such as QA and language inference.

~\citep{raffel2019} studies different pre-training objectives, architectures, unlabeled data sets, transfer approaches on various language understanding tasks. Subsequently, the authors develop pre-trained models that achieve state-of-the-art results on different tasks, i.e. summarization, QA, text classification. As part of this work, the authors also released the \href{https://www.tensorflow.org/datasets/catalog/c4}{C4} dataset which they used to train some of their pre-trained models.

In~\citep{zhang2020}, novel self-supervised objective is leveraged for pre-trained large Transformer~\citep{vaswani2017} based encoder-decoder models on massive text corpora. The authors mask important sentences from the input document, and attempt to generate the masked content from the remaining sentences. This is how extractive summarization methods typically work. The authors demonstrate the efficacy of their approach via exhaustive empirical results on a variety of summarization tasks, achieving state-of-the-art results.

Routing Transformer (RT)~\citep{roy2021} proposes to learn dynamic sparse attention patterns that avoid allocating computation and memory based resources to attend to content unrelated to the query. RT modifies the vanilla self-attention strategy with a sparse routing module based on online k-means. Thus the overall attention complexity is reduced from O($n^2d$) to O($n^{1.5}d$), given sequence length $n$ and hidden dimension $d$. The authors demonstrate the efficacy of their approach via exhaustive experimentation. RT achieves state-of-the-art results in different language modeling tasks (in particular for LFQA).

\section{Methodology}\label{sec:methodology}

As previously discussed, the aim in this work is to address all three issues impeding the progress of LFQA, as outlined in~\citep{krishna2021}. We discuss our methodology towards solving each of the above tasks in Sections~\ref{ssec:data_overlap},~\ref{ssec:automatic_metrics} and~\ref{ssec:factual_grounding}.

\subsection{Data overlap}\label{ssec:data_overlap}

Experiments confirm the ``data leakage'' issue in ELI5~\citep{fan2019}. To adequately address the train/validation/test overlap, a framework/tool is proposed, which can remove this overlap automatically. Initially, the data consisted of  \href{https://github.com/huggingface/datasets}{HF datasets}.  Currently, agglomerative hierarchical clustering (AHC)~\citep{day1984} is under evaluation for overlap removal. Developing these tools will require significant effort well beyond the time box of this project.

\subsection{Automatic metrics for LFQA tasks}\label{ssec:automatic_metrics}

For automatic metrics for LFQA tasks, adapting the CoCo~\citep{xie2021} metric was evaluated.  It was originally intended for text summarization tasks. Using the HF \href{https://huggingface.co/transformers/}{transformers} library, the summarization results outlined in~\citep{xie2021} were confirmed. Algorithm~\ref{alg:coco_lfqa} outlines the proposed CoCo metric for LFQA tasks\footnote{The code is publicly available \href{https://github.com/schrilax/xcs224u}{here}.}.

A brief overview of the proposed approach. First an LFQA retriever model $\mathcal{\boldsymbol {M}}$ is utilized to get the relevant documents $\{{\boldsymbol R}_i\}_{i = 1}^{n}$ given a question $\mathcal{\boldsymbol {Q}}$. Subsequently, the text concatenated for illustration purposes\footnote{Note that there are multiple ways to use the output from $\mathcal{\boldsymbol {M}}$ to form the source document $\mathcal{\boldsymbol {S}}$. For example, using top-{\em K} retrievals only, considering information content and overlap, ordering of results.} $\{{\boldsymbol R}_i\}_{i = 1}^{n}$ to formulate the source document $\mathcal{\boldsymbol {X}}$. Once we have the source document, we mask significant tokens i.e. for nouns, verbs, etc. (taken from the answer $\mathcal{\boldsymbol {A}}$) in $\mathcal{\boldsymbol {S}}$ to create a masked version $\mathcal{\boldsymbol {X}}^\prime$ of $\mathcal{\boldsymbol {X}}$. Next we feed both $\mathcal{\boldsymbol {X}}$ and $\mathcal{\boldsymbol {X}}^\prime$ into the scoring model $\mathcal{\boldsymbol {F}}$ and generate token probabilities. Finally we use the token probabilities to compute the CoCo score.

\begin{algorithm}[H]
  \caption{Proposed CoCo metric for LFQA}
  \label{alg:coco_lfqa}
  \begin{algorithmic}[1]
	\Require Question $\mathcal{\boldsymbol {Q}}$, LFQA retriever model $\mathcal{\boldsymbol {M}}$, generated answer $\mathcal{\boldsymbol {A}}$, scoring model $\mathcal{\boldsymbol {F}}$
	\Ensure  CoCo score $\mathcal{\boldsymbol {C}}^\prime$ of $\mathcal{\boldsymbol {A}}$ w.r.t. $\mathcal{\boldsymbol {Q}}$
	\State Get the output $\{{\boldsymbol R}_i\}_{i = 1}^{n}$ of $\mathcal{\boldsymbol {M}}$ given $\mathcal{\boldsymbol {Q}}$.
	\State Concatenate $\{{\boldsymbol R}_i\}_{i = 1}^{n}$ to form the source document $\mathcal{\boldsymbol {X}}$.
	\State Select key tokens $\mathcal{\boldsymbol {A}}^\prime$ from $\mathcal{\boldsymbol {A}}$.
	\State Mask the source document $\mathcal{\boldsymbol {X}}$ according to $\mathcal{\boldsymbol {A}}^\prime$ to produce $\mathcal{\boldsymbol {X}}^\prime$.
	\State Feed $\mathcal{\boldsymbol {X}}$ and $\mathcal{\boldsymbol {X}}^\prime$ into scoring model $\mathcal{\boldsymbol {F}}$ to generate token probabilities, i.e., $P (\boldsymbol{a}_i | \mathcal{\boldsymbol {X}}, \boldsymbol{a}_{< i})$ and $P (\boldsymbol{a}_i | \mathcal{\boldsymbol {X}}^\prime, \boldsymbol{a}_{< i})$, $\forall \boldsymbol{a}_i \in \mathcal{\boldsymbol {A}}^\prime$.
	\State Calculate $\mathcal{\boldsymbol {C}}^\prime = \frac{1}{| \mathcal{\boldsymbol {A}}^\prime |} \sum\limits_{\boldsymbol{a}_t \in \mathcal{\boldsymbol {A}}^\prime} P (\boldsymbol{a}_t | \mathcal{\boldsymbol {X}}, \boldsymbol{a}_{< t}) - P (\boldsymbol{a}_t | \mathcal{\boldsymbol {X}}^\prime, \boldsymbol{a}_{< t})$
    \end{algorithmic}
\end{algorithm}

\subsection{Factual grounding of answers}\label{ssec:factual_grounding}

The CoCo score is applied to measure how grounded generated answers are with respect to retrieved documents. Algorithm~\ref{alg:coco_grounding} outlines our approach. By comparing the CoCo score from both top-{\em K} retrievals from the LFQA retriever model $\mathcal{\boldsymbol {M}}$ and random retrievals, we can quantitatively measure how ``grounded'' generated answers are to the retrieved documents.

\begin{algorithm}[H]
  \caption{Using our proposed CoCo metric to measure grounding of answers}
  \label{alg:coco_grounding}
  \begin{algorithmic}[1]
	\Require Question $\mathcal{\boldsymbol {Q}}$, LFQA retriever model $\mathcal{\boldsymbol {M}}$, generated answer $\mathcal{\boldsymbol {A}}$, scoring model $\mathcal{\boldsymbol {F}}$
	\Ensure  Grounding score $\mathcal{\boldsymbol {G}}$ of $\mathcal{\boldsymbol {A}}$ w.r.t. $\mathcal{\boldsymbol {Q}}$
	\State Compute the CoCo score $\mathcal{\boldsymbol {C}}_{\text{top-K}}$ using top-{\em K} retrievals using LFQA retriever model $\boldsymbol M$.
	\State Compute the CoCo score $\mathcal{\boldsymbol {C}}_{\text{random}}$ using random retrievals.
	\State Compute $\mathcal{\boldsymbol {G}} = $  $\mathcal{\boldsymbol {C}}_{\text{top-K}} - \mathcal{\boldsymbol {C}}_{\text{random}}$.
    \end{algorithmic}
\end{algorithm}

Thus this work addresses all the issues identified in~\citep{krishna2021}, via novel application of the CoCo metric for LFQA tasks.

\section{Results}\label{sec:results}
This section exemplifies the performance of the proposed metric on baseline datasets. As we highlighted earlier, the goal of this work is to address all three issues impeding the progress of LFQA, as outlined in~\citep{krishna2021}. The highlighted experiments illustrate the extent to which these bottlenecks are ameliorated. The experimental setup is described in Section~\ref{ssec:experimental_setup}, including experiments to mitigating the three issues highlighted above. Apart from the experiments in Sections~\ref{ssec:data_overlap_results},~\ref{ssec:automatic_metrics_results} and~\ref{ssec:factual_grounding_results}, note Section~\ref{sec:appendix} that contains additional experiments, which we could not include in the main text due to space constraints.

\subsection{Experimental setup}\label{ssec:experimental_setup}

The experiments rely on the original~\citep{fan2019} and  \href{https://huggingface.co/datasets/vblagoje/eli5}{new} ELI5 datasets. The original ELI5 dataset is significant since it was the earliest large-scale corpus for LFQA tasks, and was instrumental in the rapid strides of progress in this field. It was made up from 270K threads from the Reddit forum “Explain Like I’m Five” (ELI5) and provides detailed and in-depth answers to open-ended questions. Compared to earlier datasets, it consists of diverse questions that require multi-sentence answers grounded on multiple texts. However this dataset suffers from ``data leakage'' issues stemming from train/validation/test question overlap.

The vanilla Sentence BERT model is applied to confirm the data overlap in the original ELI5 dataset. For the other experiments, BART based models is relied upon extensively for token based scoring, as part of our proposed CoCo metric for LFQA tasks.

\subsection{Data overlap}\label{ssec:data_overlap_results}

The Sentence-BERT (SBERT)~\citep{reimers2019} semantic search utility and the HuggingFace (HF) \href{https://huggingface.co/datasets/eli5}{ELI5} dataset are applied to confirm the ``data leakage'' issue. This is accomplised via an all-to-all comparison of ELI5 questions, using the cosine similarity metric. More specifically, questions in the train, test and validation sets are compared to gauge semantic similarity (relying on their neural embeddings). Indeed, there is a significant semantic overlap between the questions in these two sets, as the results show. Note that we compare top-{\em K} similarity scores (for {\em K} = $1, 2, 3$) for each question. A good portion of the questions are both test/validation sets, corresponding ``paraphrased'' questions are part of the datasets.

\begin{figure}[!htbp]
\centering
\includegraphics[scale=0.45]{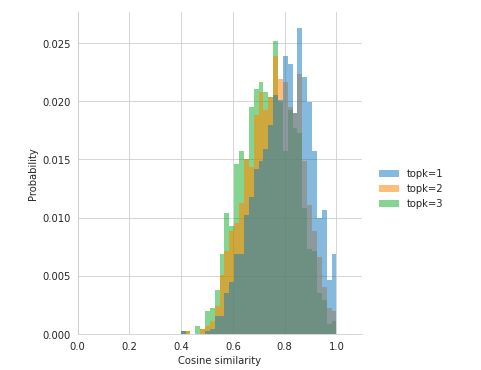}
\caption{Train/validation dataset overlap in the ELI5~\citep{fan2019} dataset via comparing top-{\em K} similarity scores. We can observe significant overlap in the probability distributions.}
\label{fig:old_eli5_train_validation}
\end{figure}

The semantic overlap in the training and validation sets is much larger than the overlap between the training and validation sets. Thus the results in~\citep{krishna2021} are confirmed, though they reported much higher overlap percentage than observed here.

\begin{figure}[!htbp]
\centering
\includegraphics[scale=0.45]{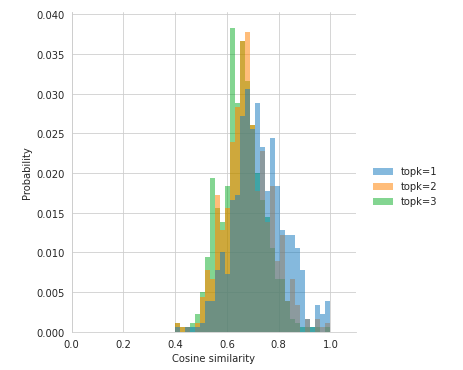}
\caption{Train/test dataset overlap in the ELI5~\citep{fan2019} dataset via comparing top-{\em K} similarity scores. We can observe significant overlap in the probability distributions.}
\label{fig:old_eli5_train_test}
\end{figure}

\begin{figure}[!htbp]
\centering
\includegraphics[scale=0.45]{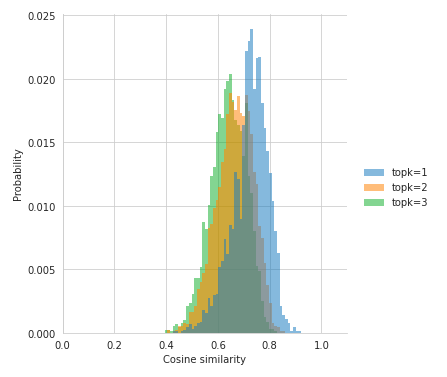}
\caption{Train/validation overlap in the new \href{https://huggingface.co/datasets/vblagoje/eli5}{ELI5} dataset via comparing top-{\em K} similarity scores. In comparison to Figure~\ref{fig:old_eli5_train_validation}, a substantial decrease in the overlap in the probability distributions is observed.}
\label{fig:eli5_train_validation}
\end{figure}

\begin{figure}[!htbp]
\centering
\includegraphics[scale=0.45]{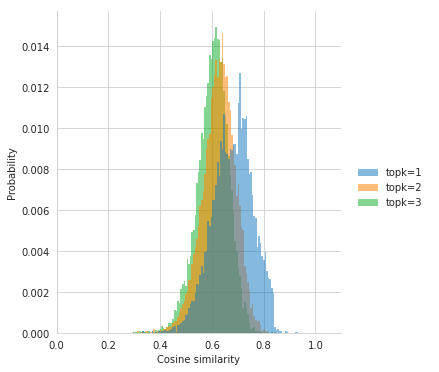}
\caption{Train/test overlap in the new \href{https://huggingface.co/datasets/vblagoje/eli5}{ELI5} dataset via comparing top-{\em K} similarity scores. In comparison to Figure~\ref{fig:old_eli5_train_test}, a substantial decrease in the overlap in the probability distributions is observed.}
\label{fig:eli5_train_test}
\end{figure}

As part of this work, the authors of~\citep{fan2019} made available a new version of the dataset, currently in alpha under preparation, but already publicly available \href{https://huggingface.co/datasets/vblagoje/eli5}{here}. In this dataset, paraphrased questions are largely removed, up to a certain threshold of similarity. As confirmed in our experiments, the new ELI5 dataset has a much lower overlap between train and validation sets. We refer the readers to Figures~\ref{fig:old_eli5_train_validation},~\ref{fig:old_eli5_train_test},~\ref{fig:eli5_train_validation} and~\ref{fig:eli5_train_test} for more details.

%as well as train/test sets.

\subsection{Automatic metrics for LFQA tasks}\label{ssec:automatic_metrics_results}

Applying the proposed CoCo score it can be observed that generated answers remain the same, irrespective of the documents identified by the LFQA retriever model. This confirms what~\citep{krishna2019} showcase via their empirical results. For this task, we utilized BART based ELI5 Seq2Seq models. However more exhaustive experimentation is required before confirming the benefits of the proposed CoCo metric.

% As we are still in the early stages of preparing a complete experiment running Coco scores for both best/worst passages and the current Bart ELI5 and the upcoming Bart ELI5 seq2seq models, we would refrain from making a conclusion on the usefulness of the Coco score for LFQA task at this time. Still, early indicators confirm what the Hurdles paper authors found - retrieved documents do not matter much for the generated answers.  

\subsection{Factual grounding of answers}\label{ssec:factual_grounding_results}

As discussed in Section~\ref{ssec:factual_grounding}, we use the proposed CoCo metric to determine how ``grounded'' generated answers are with respect to the retrieved documents in a LFQA system. Consider as example the question ``Why does water heated to room temperature feel colder than the air around it ?'', we follow the procedure outlined in Algorithm~\ref{alg:coco_grounding}. Subsequently we computed the CoCo scores for the following scenarios: i) top-{\em K} passages output by the LFQA retriever model, as well as ii) random passages unrelated to the question mentioned above. The results we got were along expected lines - the CoCo score for the unrelated random passages was actually better than that using the top-{\em K} passages by 0.147. This confirms the hypothesis outlined by~\citep{krishna2019}, i.e., generated answers do not seem to be ``grounded'' on conditioned passages but are most likely intrinsic to the language model itself.

% We refer the readers to Algorithm~\ref{alg:coco_grounding} for more details

%Finally, we moved on to the factual grounding of answers experiments. We took the retriever best passages for the "Why does water heated to room temperature feel colder than the air around it?" question, computed its Coco score and compared it to Coco score for random passages unrelated to the question mentioned above. The results were not surprising - the Coco score for the unrelated random passages was better than for the best passages - 0.147. This result further confirms the hypothesis made by Hurdles authors - generated answers are not grounded on conditioned passages but are most likely intrinsic to the language model itself. 

\section{Conclusion}\label{sec:conclusion}

This work verifies the issues raised in~\citep{krishna2019}. Firstly using Sentence BERT, the significant overlap that exists between train/validation questions in the ELI5~\citep{fan2019} dataset is confirmed. This is the only large-scale publicly available dataset for LFQA tasks, critical for community progress. As experimented, the upcoming alpha ELI5 version, combined with the new Seq2Seq BART ELI5 model, are likely to greatly mitigate the overlap issue.

Secondly, a novel application of the CoCo metric is proposed, to evaluate how factual answers are to questions.  Experimental results show promising results for its usage as an automatic metric for LFQA tasks. Additional exhaustive experimentation is required to confirm its utility. These experiments could be extended to compare our CoCo scores for all test questions, along the best/worst retrieved passages and new/old BART ELI5 Seq2Seq model options. If the CoCo scores indeed improve for the new BART ELI5 using best passages, while simultaneously remain low for the other three possible options, we would be much more optimistic in the correctness and usefulness of the CoCo score for LFQA. It is worth mentioning the CoCo itself alone should not be an automatic choice for the LFQA, as it only measures one dimension: factuality. Additional metrics will be needed to evaluate dimensions like coherence and fluency as well.

Thirdly, a novel application of the CoCo metric is proposed to measure the ``grounding'' of generated answers based on the retrieved documents. Thus every one of the important issues raised in~\citep{krishna2019} is addressed.  

% \bibliographystyle{acl_natbib}
% \bibliography{anthology,acl2021}

\appendix
\section{Appendix}\label{sec:appendix}
\begin{figure}[!htbp]
\centering
\includegraphics[scale=0.45]{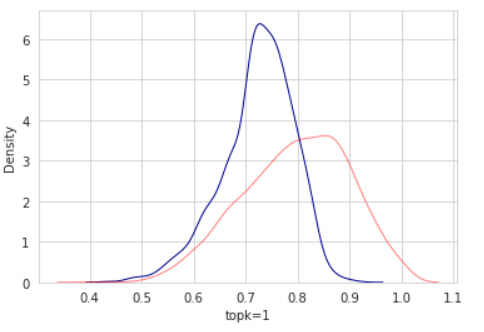}
\caption{Comparing the density between top-{\em K} similarity scores for $k = 1$ between old (shown in red) and new (shown in blue) datasets for training/validation. There is a significant reduction in similarity scores.}
\label{fig:topk1}
\end{figure}

\begin{figure}[!htbp]
\centering
\includegraphics[scale=0.45]{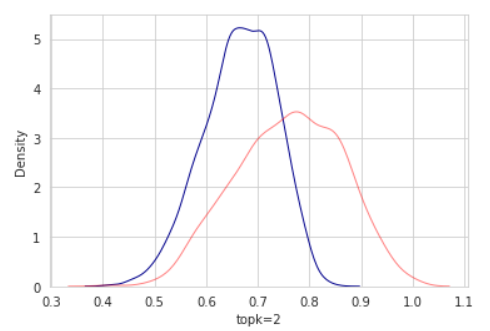}
\caption{Comparing the density between top-{\em K} similarity scores for $k = 2$ between old (shown in red) and new (shown in blue) ELI5 datasets for training/validation. There is a significant reduction in similarity scores.}
\label{fig:topk2}
\end{figure}

\begin{figure}[!htbp]
\centering
\includegraphics[scale=0.45]{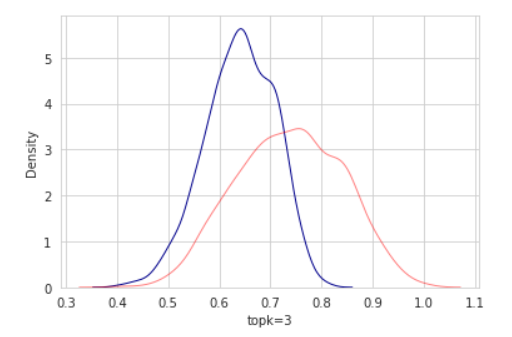}
\caption{Comparing the density between top-{\em K} similarity scores for $k = 3$ between between old (shown in red) and new (shown in blue) datasets for training/validation. There is a significant reduction in similarity scores.}
\label{fig:topk3}
\end{figure}

This section includes additional results. Specifically, more fine-grained analysis of the training/validation dataset overlap between~\citep{fan2019} and \href{https://huggingface.co/datasets/vblagoje/eli5}{new} ELI5 datasets. Figures~\ref{fig:topk1},~\ref{fig:topk2} and~\ref{fig:topk3} contrast the differences in detail. It can be clearly observed  by the similarity score distribution, which has shifted to the left for the new alpha dataset, in each of the results. Experiments indicate there is a 13\% reduction in similarity scores and given the similarity scores directly affects the ``overlap''.  Thus, the new dataset has ameliorated the ``data leakage'' issue significantly. Kurtosis scores have also improved in the new ELI5 dataset, given the distribution of scores are also closer to the mean and are more peaked, thus increasing our confidence index.

\end{document}